\documentclass[journal]{IEEEtran}
\usepackage{graphicx}
\usepackage{multirow}
\usepackage{multicol}
\usepackage{amsmath,amssymb,amsfonts}
\usepackage{amsthm}
\usepackage{mathrsfs}
\usepackage{xcolor}
\usepackage{textcomp}
\usepackage{manyfoot}
\usepackage{booktabs}
\usepackage{algorithm}
\usepackage{algorithmicx}
\usepackage{algpseudocode}
\usepackage{listings}
\usepackage{ifpdf}
\usepackage{array}
\usepackage{url}
\usepackage{subcaption}
\usepackage{caption}
\usepackage{tabularray}
\usepackage{tikz}
\usepackage{float}
\usepackage{hyperref}
\usepackage{colortbl}
\usepackage{lscape}
\usepackage{rotating}
\usepackage{soul}
\usepackage{natbib}
\usepackage{todonotes}
\usepackage{etoolbox}
\usepackage{enumitem}
\usepackage[utf8]{inputenc}
\usepackage{pifont}
\usepackage{balance}

\newcommand{\cmark}{\ding{51}}
\newcommand{\xmark}{\ding{55}}

\begin{document}

\title{FR-GESTURE: An RGBD Dataset For Gesture-based Human-Robot Interaction In First Responder Operations}

\author{
    \IEEEauthorblockN{
        Konstantinos~Foteinos\IEEEauthorrefmark{1},
        Georgios~Angelidis\IEEEauthorrefmark{1},
        Aggelos~Psiris\IEEEauthorrefmark{1},
        Vasileios~Argyriou\IEEEauthorrefmark{2},
        Panagiotis~Sarigiannidis\IEEEauthorrefmark{3},
        Georgios~Th.~Papadopoulos\IEEEauthorrefmark{1}\IEEEauthorrefmark{4}
    }

    \IEEEauthorblockA{
        \IEEEauthorrefmark{1}
        Department of Informatics and Telematics,
        Harokopio University of Athens, Athens, Greece
    }

    \IEEEauthorblockA{
        \IEEEauthorrefmark{2}
        Department of Networks and Digital Media, 
        Kingston University, Kingston upon Thames, Surrey, United Kingdom
    }

    \IEEEauthorblockA{
        \IEEEauthorrefmark{3}
        Department of Electrical and Computer Engineering, 
        University of Western Macedonia, Kozani, Greece
    }

    \IEEEauthorblockA{
        \IEEEauthorrefmark{4}
        Archimedes, Athena Research Center, Athens, Greece\\
    }

    \IEEEauthorblockA{
        \IEEEauthorrefmark{1}\{kfoteinos,gangelidis,aggelospsiris,g.th.papadopoulos\}@hua.gr
        \IEEEauthorrefmark{2}vasileios.argyriou@kingston.ac.uk
        \IEEEauthorrefmark{3}psarigiannidis@uowm.gr
    }

    \thanks{This work has received funding from the European Union’s Horizon Europe research and innovation programme under Grant Agreement No. 101168042 project TRIFFID (auTonomous Robotic aId For increasing First responders Efficiency). The views and opinions expressed in this paper are those of the authors only and do not necessarily reflect those of the European Union or the European Commission.}
    
}

\maketitle

\begin{abstract}
    The ever increasing intensity and number of disasters make even more difficult the work of First Responders (FRs). Artificial intelligence and robotics solutions could facilitate their operations, compensating these difficulties. To this end, we propose a dataset for gesture-based UGV control by FRs, introducing a set of 12 commands, drawing inspiration from existing gestures used by FRs and tactical hand signals and refined after incorporating feedback from experienced FRs. Then we proceed with the data collection itself, resulting in 3312 RGBD pairs captured from 2 viewpoints and 7 distances. To the best of our knowledge, this is the first dataset especially intended for gesture-based UGV guidance by FRs. Finally we define evaluation protocols for our RGBD dataset, termed FR-GESTURE, and we perform baseline experiments, which are put forward for improvement. We have made data publicly available to promote future research on the domain: \href{https://doi.org/10.5281/zenodo.18131333}{https://doi.org/10.5281/zenodo.18131333}.
\end{abstract}


\IEEEpeerreviewmaketitle

\section{Introduction}

    \IEEEPARstart{G}{esture} recognition \cite{linardakis2025survey} plays a crucial role in various applications domains, including Human-Robot Interaction (HRI), Human-Computer Interaction (HCI) and Sign Language Understanding (SLU) \cite{papadopoulos2022user, moutousi2025tornado, papadopoulos2021towards}. The rapid evolution of Deep Learning (DL) \cite{konstantakos2025self, cani2026illicit, evangelatos2025exploring} and the ever-increasing size of datasets has shifted the research community to study more complicated tasks that cannot be solved by traditional Computer Vision (CV) techniques, such as ultra-range gesture classification \cite{urgr_bamani2024ultra} and continuous Sign Language Recognition (SLR) \cite{vac_cslr_min2021visual}. However, several challenges remain: data scarcity, computational cost, presence of gesture-irrelevant factors and limited capacity of the existing architecture are issues that have not been solved yet \cite{foteinos2025visual}.
    
    Gesture recognition approaches can be divided in two categories: sensor-based and vision-based. Sensor-based techniques rely on devices such as wearable sensors (e.g. glove-based) or non-wearable (e.g. using acoustic signals). On the contrary, Visual Hand Gesture Recognition (VHGR) leverages RGB, depth and derived modalities, like skeleton and optical flow, either estimated or captured by specialized cameras. Even though sensor-based approaches have access to less redundant and more accurate information, the expensiveness of deploying these - often cumbersome - sensors limit their applicability in real-world scenarios. Thus, VHGR has dominated over sensor-based, being more suitable for real-world deployment. Convolutional Neural Networks (CNNs) are mainly utilized for unimodal appearance-based \cite{gesture_cnn_sharma2021vision} and GCNs for solely pose-based \cite{td_gcn_liu2023temporal} VHGR. Multimodal VHGR approaches combine these modalities often in multi-stream settings \cite{ctfb_hampiholi2023convolutional}. Multimodal gesture recognition, employing both sensor and visual data, has also been examined by the research community.

    In the context of gesture-based Autonomous Mobile Robot (AMR) guidance, VHGR seems more suitable approach as it relies only on camera input, which is usually already mounted on the AMR for other purposes. Furthermore, as discussed earlier, intrusive sensors (e.g. gloves) may reduce the ease of use becoming suboptimal means of HRI. In this paper, we consider AMRs \cite{linardakis2024distributed} used by First Responders (FRs) for enhancing situational awareness \cite{TRIFFID-cani2025triffid} and providing useful functionalities (e.g. fetching rescue tools). As it is not feasible to equip all FRs with devices (e.g. remote controllers) to give commands to the AMRs and this would distract them and interrupt their operations, VHGR could constitute a more convenient and intuitive form of HRI in these scenarios. The current paper aims to address this, by introducing the first, to the best of our knowledge, hand signal dataset for gesture-based AMR control by FRs. Existing datasets for AMR guidance via gestures even focus on UAV navigation recognizing hand signals for aircraft marshaling \cite{NATOPS_5771448}, consider gestures borrowed from sign languages \cite{LRHG_9561189} or comprise only a limited number of commands \cite{urgr_bamani2024ultra}. Moreover, only a few of them are publicly available.

    To this end, we propose 12 gestures mapped to UGV commands to facilitate gesture-based AMR guidance, tailored for first response operations. In particular, we have collected and made publicly available 3312 RGBD pairs of gestures captured at 7 different distances to increase robustness to recognition range variability. Further, the dataset was gathered in 3 distinct environments (2 indoors and 1 outdoors) in order to make the dataset more diverse. These samples are used to train image classifiers, following the two introduced evaluation protocols (i.e., uniform and subject independent).

    In summary, the contributions of our study are the following:
    \begin{itemize}
        \item \textbf{A novel corpus for gesture-based HRI:} Motivated by the absence of a complete set of commands for HRI, we proceed with the definition of a gesture/command mapping, especially according to FRs needs.
        \item \textbf{A public dataset for static VHGR:} Based on the aforementioned corpus, we have collected an RGBD dataset, termed FR-GESTURE, which is made publicly available to the research community \footnote{https://doi.org/10.5281/zenodo.18131332}.
        \item \textbf{Baseline experiments:} Finally, we define evaluation protocols and execute basic experiments to establish a baseline for future work.
    \end{itemize}

    The rest of the paper is structured as follows. Section \ref{sec:related-work} discusses related work on VHGR, image classification and gesture-based AMR guidance. Section \ref{sec:dataset} introduces the custom corpus and the FR-GESTURE dataset. Section \ref{sec:experiments} presents the experimental setup and results. Section \ref{sec:limitations} highlights limitations of the FR-DATASET making suggestions for future work and Section \ref{sec:conclusion} concludes the study.
    
\section{Related work}
    \label{sec:related-work}

    As mentioned in the previous section, VHGR has been thoroughly investigated by the research community. In the next subsections, relevant approaches are examined, while a brief introduction to image classification is also conducted, regarding its strong relevance to static VHGR. Finally, existing datasets for gesture-based AMR control are analyzed.

    \subsection{Methods for VHGR}

    In general, VHGR methods can be categorized into dynamic (DHGR) and static (SHGR), based on whether they consider the temporal domain or not. Dynamic approaches are more suitable for classifying more complicated, moving gestures, especially in the context of SLR, but require more computational resources. On the contrary, static methods utilize only spatial information, failing to capture hand movements but recognizing gestures at lower latency.

    In more detail, methods for SHGR are in their overwhelming majority RGB image classifiers, obtained either by adopting existing general-purpose architectures or through custom configuration. For static SLR, both 2DCNNs \cite{gesture_cnn_sharma2021vision} and transformers \cite{signformer_kothadiya2023signformer} have been examined, achieving high accuracy even with simple architectures. Approaches for HCI focus on computational efficiency in order to enable real-time recognition \cite{lhgr_net_zhang2023lightweight}, often applying hand palm segmentation to remove background interference \cite{fgds_net_zhou2024fgdsnet,pipelines_dang2023lightweight}. Finally, for HRI, guiding UGVs through gestures in long distances were considered by authors in \cite{urgr_bamani2024ultra}, who use human detection to normalize the spatial dimensions and super resolution to improve the quality of the cropped patch.

    Methods for DHGR can be further divided into isolated (IDHGR) and continuous (CDHGR), considering a single or multiple gesture instances per input sequence, respectively. For IDHGR, various visual modalities have been examined more extensively compared to SHGR, including pose and optical flow, as well as strategies to fuse them \cite{mmtm_joze2020mmtm,ctfb_hampiholi2023convolutional}. In particular, for unimodal pose-based IDHGR, ST-GCN variants \cite{td_gcn_liu2023temporal,dstsa_gcn_cui2025dstsa} have been widely adopted and GCN+Transformer architectures \cite{signbert_plus_hu2023signbert+,best_zhao2023best} have been studied thoroughly in the context of BERT-inspired pretraining. Regarding limitations of the skeleton modality (i.e. due to pose estimation failures), pose-based approaches are often combined with RGB-based 3DCNNs resulting in dual-stream setups. RGB-based CDHGR methods have dominated over unimodal pose-based, as the pose estimation fails in the latter case due to the fast hand movements. The architectures follow mainly the 2DCNN+1DCNN+BiLSTM pipeline scheme with extra supervision \cite{vac_cslr_min2021visual}. Modifications include replacing the 2DCNN with Foundation Models (FMs) \cite{clip_sla_alyami2025clip} and using diffusion models \cite{cslr_dm_geng2026no} for sequence generation. Multimodal approaches for CHGR have been less studied, with an indicative exampled being \cite{two_stream_slr_chen2022two} that achieves superior performance.
    
    \subsection{Image classification}

    SHGR can be viewed as a specialization of the general-purpose image classification task \cite{rodis2024multimodal}. Indeed, many such methods have been proposed for SHGR \cite{signformer_kothadiya2023signformer,kapitanov2024hagrid,pipelines_dang2023lightweight} or utilized as spatial perception modules \cite{vac_cslr_min2021visual} and proved effective.
    
    CNNs were the first type of DL image classifiers proposed by the researchers, based on learnable convolution filters \cite{mademlis2024invisible}. ResNet \cite{resnet_he2016deep} and its variants \cite{resnext_xie2017aggregated} increased performance by introducing residual connections, facilitating gradient flow. More recent approaches focused on computational complexity, either using depthwise separable convolution \cite{mobilenet_howard2017mobilenets} or determining optimal ways to adjust networks size \cite{efficient_net_tan2019efficientnet}. The current state-of-the-art is engaged with increasing the network capacity maintaining the computational cost \cite{parameter_net_han2024parameternet} and making a step forward to FMs \cite{intern_image_wang2023internimage}.

    Transformers were originally proposed for Natural Language Processing (NLP) tasks. However, they were adopted for CV by authors in \cite{vit_dosovitskiy2020image}, who introduced a fully attention-based approach, which treats images as patches sequence and combines linear embeddings with position encoding. Later approaches have been based on this, aiming to reduce the computational cost through non-overlapped windows \cite{swin_t_liu2021swin} or to enhance it via knowledge distillation \cite{deit_touvron2021training}. Hybrid architectures, which combine the global receptive fields of the transformers with the locality of convolutions have also been proposed \cite{flatten_han2023flatten}.

    More advanced techniques have been recently introduced, including the spatial (or visual) Multi-Layer Perceptrons (MLPs), which incorporate the same architecture with transformers but replace attention with FC layers \cite{mlp_mixer_tolstikhin2021mlp}. The capability of Graph Neural Networks (GNNs) to capture fine-grained spatial relationships was investigated by the authors in \cite{vision_gnn_han2022vision}, treating image patches as nodes. Finally, State Space Modes (SSMs) were recently considered for NLP and CV, aiming to overcome transformers' limitations. Mamba variants are often deployed for this purpose \cite{vmamba_liu2024vmamba}.
    
    \subsection{VHGR datasets for AMR guidance}

    Only a few datasets have been introduced for AMR guidance and they are reported in the Table \ref{tab:amr-hgr-datasets}. URGR \cite{urgr_bamani2024ultra} was the only dataset for controlling UGVs and it is the largest one in the list, comprising almost 350K samples performed by 16 signers. Having been captured in various and large distances (1-25m) in both indoor and outdoor environments, it is suitable for real-world applications. However, the limited number of commands (only 5) restricts the robot capabilities. MD-UHGRD \cite{MD_UHGRD_10651436} contains 19 fine-grained gestures/commands (and a no-gesture class) related to UAV navigation and functionalities and it was captured with a camera mounted on a drone, at heights 1.8-3m and distances 3-10m. LRHG \cite{LRHG_9561189} was proposed also for UAV navigation, but its 10 gestures are borrowed from a sign language. However, it has been collected by three more participants, improving the generalization ability to unseen subjects, compared to MD-UHGRD. Both of them provide bounding boxes annotations. The authors in \cite{construction_WANG2021103625} proposed 11 commands for navigating autonomous construction machinery relying on RGBD videos. The AUTH UAV GESTURE \cite{auth_uav_9571027} comprises 6 gestures, following NATOPS \footnote{https://studylib.net/doc/8722364/navair-00-80t-113}, performed by 20 subjects. These samples are combined with corresponding videos from the NTU \cite{ntu_7780484} dataset. NATOPS hand signals were also considered in UAV GESTURE \cite{UAV-GESTURE} and NATOPS \cite{NATOPS_5771448}, recorded outdoors and indoors, respectively.

    \begin{table*}
        \centering
        \caption{Existing datasets for gesture-based AMR guidance.}
        \setlength{\tabcolsep}{3pt}
        \renewcommand{\arraystretch}{1.2}
        \begin{tabular}{
            lcccccccccc
        }
            \toprule
            \textbf{Name} & \textbf{Year} & \textbf{Task} & \textbf{Robot} & \textbf{Classes} & \textbf{Samples} & \textbf{Subjects} & \textbf{Environment} & \textbf{Range} & \textbf{Modality} & \textbf{Public} \\
            \midrule
            URGR \cite{urgr_bamani2024ultra} & 2024 & SHGR & UGV & 5 & 347483 & 16 & Indoor \& outdoor & 1-25 m & RGB & \xmark \\ 
            MD-UHGRD \cite{MD_UHGRD_10651436} & 2024 & SHGR & UAV & 19+1 & 20000 & 5 & - & 3-10 m & RGB & \xmark \\ 
            LRHG \cite{LRHG_9561189} & 2021 & SHGR & UAV & 10 & 4320 & 8 & Indoor & 1-7 m & RGB & \cmark \\ 
            \cite{construction_WANG2021103625} & 2021 & IDHGR & Construction Machinery & 11 & 364 & - & Indoor \& outdoor & - & Video RGBD & \xmark \\ 
            AUTH \cite{auth_uav_9571027} & 2021 & IDHGR & UAV & 6 & 4930 & 20 & Indoor \& outdoor & - & Video RGB & \xmark \\ 
            UAV GESTURE \cite{UAV-GESTURE} & 2019 & IDHGR & UAV & 13 & 119 & 8 & Outdoor & - & Video RGB & \xmark \\ 
            NATOPS \cite{NATOPS_5771448} & 2011 & IDHGR & UAV & 24 & 9600 & 20 & Indoor & - & Video RGBD & \cmark \\
            \midrule
            FR-GESTURE & 2026 & SHGR & UGV & 12 & 3312 & 7 & Indoor \& outdoor & 1-7 m & RGBD & \cmark \\
            \bottomrule
        \end{tabular}
        \label{tab:amr-hgr-datasets}
    \end{table*}

\section{Dataset}
    \label{sec:dataset}

\begin{figure}
        \centering
        \begin{tabular}{ccc}
             \begin{subfigure}[t]{0.13\textwidth}
                \centering
                \includegraphics[width=\linewidth]{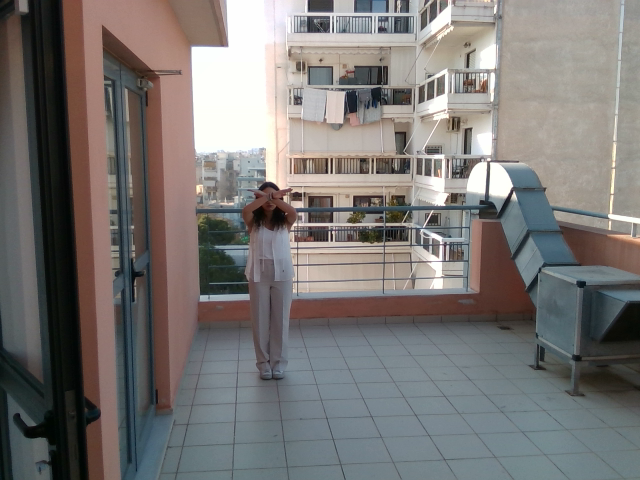}
            \end{subfigure}&
            \begin{subfigure}[t]{0.13\textwidth}
                \centering
                \includegraphics[width=\linewidth]{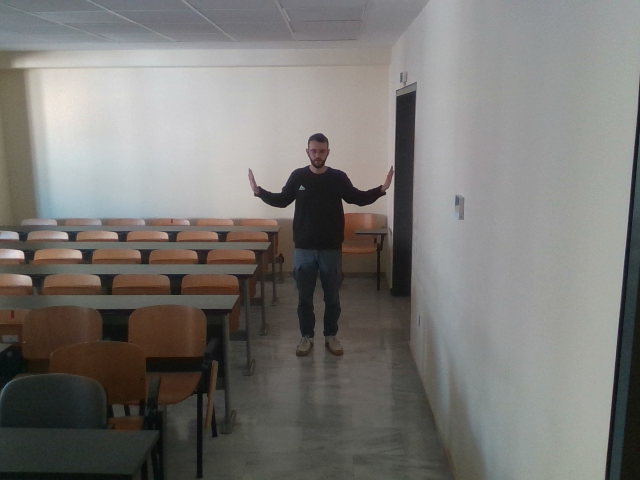}
            \end{subfigure}&
            \begin{subfigure}[t]{0.13\textwidth}
                \centering
                \includegraphics[width=\linewidth]{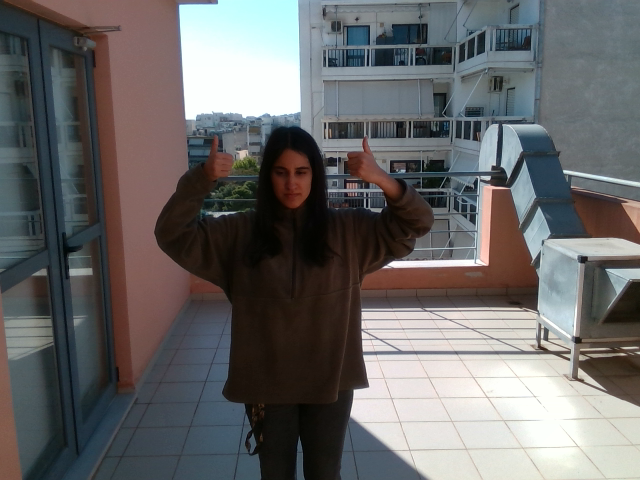}
            \end{subfigure}\\
            \begin{subfigure}[t]{0.13\textwidth}
                \centering
                \includegraphics[width=\linewidth]{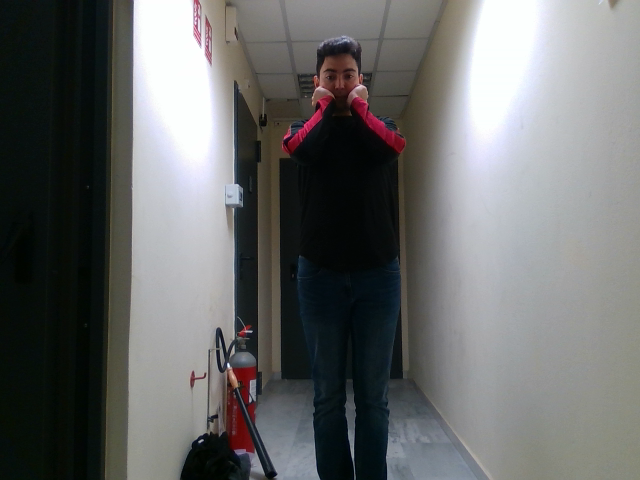}
            \end{subfigure}&
            \begin{subfigure}[t]{0.13\textwidth}
                \centering
                \includegraphics[width=\linewidth]{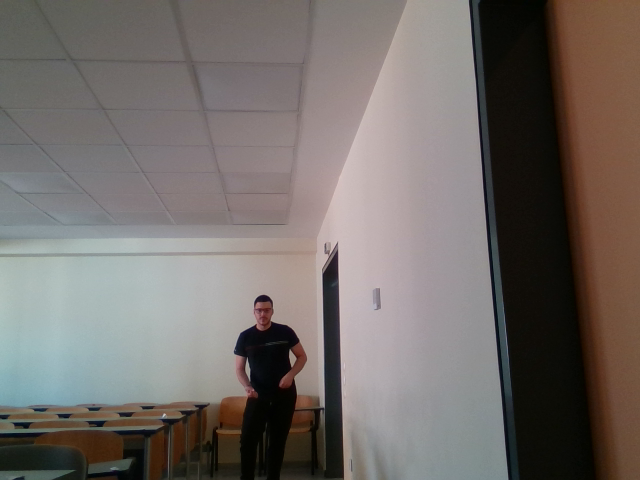}
            \end{subfigure}&
            \begin{subfigure}[t]{0.13\textwidth}
                \centering
                \includegraphics[width=\linewidth]{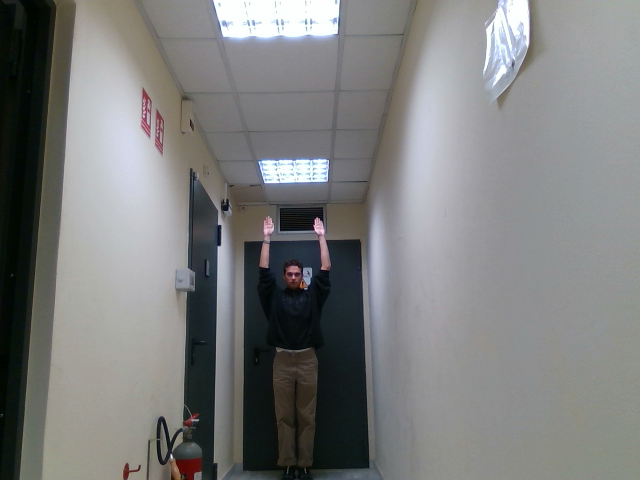}
            \end{subfigure}\\
            \begin{subfigure}[t]{0.13\textwidth}
                \centering
                \includegraphics[width=\linewidth]{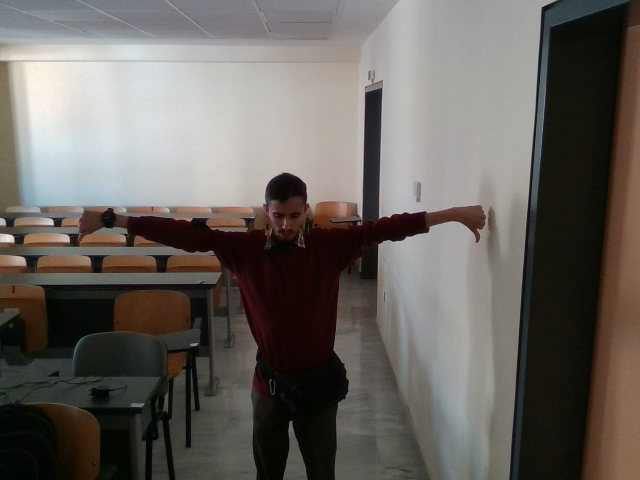}
            \end{subfigure}&
            \begin{subfigure}[t]{0.13\textwidth}
                \centering
                \includegraphics[width=\linewidth]{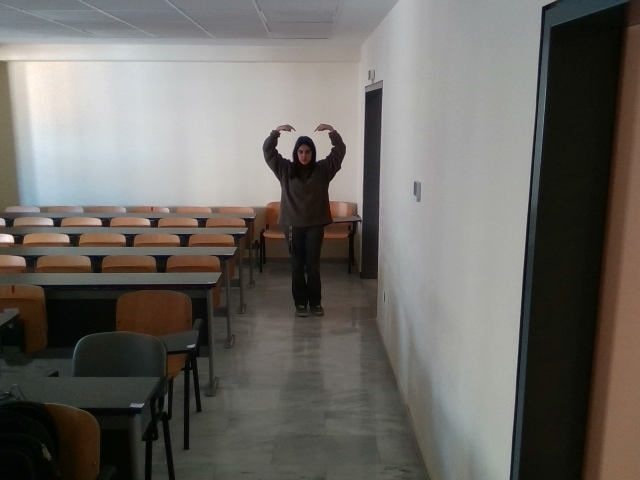}
            \end{subfigure}&
            \begin{subfigure}[t]{0.13\textwidth}
                \centering
                \includegraphics[width=\linewidth]{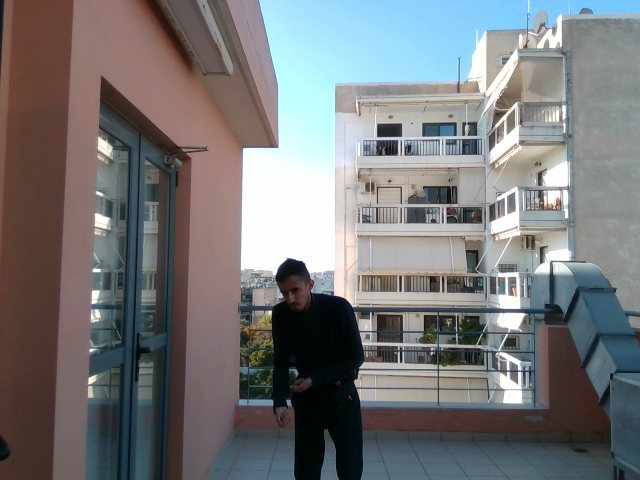}
            \end{subfigure}\\
            \begin{subfigure}[t]{0.13\textwidth}
                \centering
                \includegraphics[width=\linewidth]{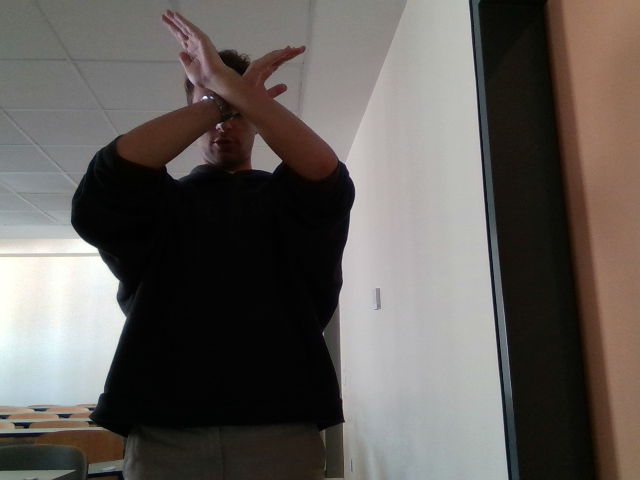}
            \end{subfigure}&
            \begin{subfigure}[t]{0.13\textwidth}
                \centering
                \includegraphics[width=\linewidth]{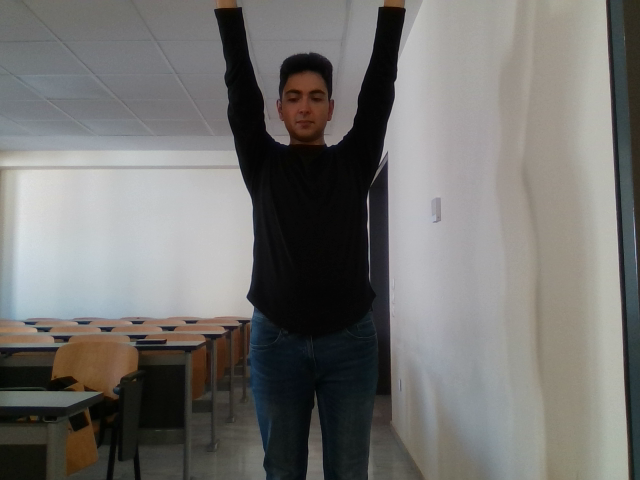}
            \end{subfigure}&
            \begin{subfigure}[t]{0.13\textwidth}
                \centering
                \includegraphics[width=\linewidth]{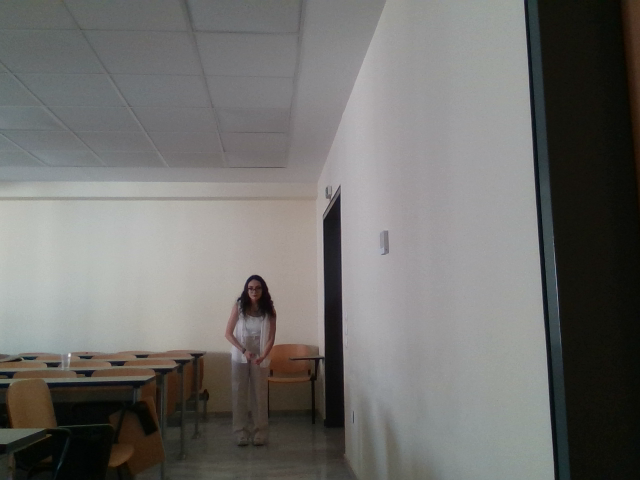}
            \end{subfigure}\\
        \end{tabular}
        \caption{Indicative samples from the FR-GESTURE dataset. Each gesture instance is captured by two cameras, positioned at different heights to increase diversity. Participants performed signs in various scenes to enhance generalization ability.}
        \label{fig:indicate-samples}
    \end{figure}

    In this section, details of the hand signals and the corresponding robot commands, the data curating procedure for collecting the FR-GESTURE dataset and related statistics are provided. Indicative samples are shown in Fig. \ref{fig:indicate-samples}.

    \subsection{Corpus definition}

\begin{figure*}[t]
        \centering
        \begin{tabular}{cccc}
            
            \begin{subfigure}[t]{0.2\textwidth}
                \centering
                \includegraphics[width=\linewidth]{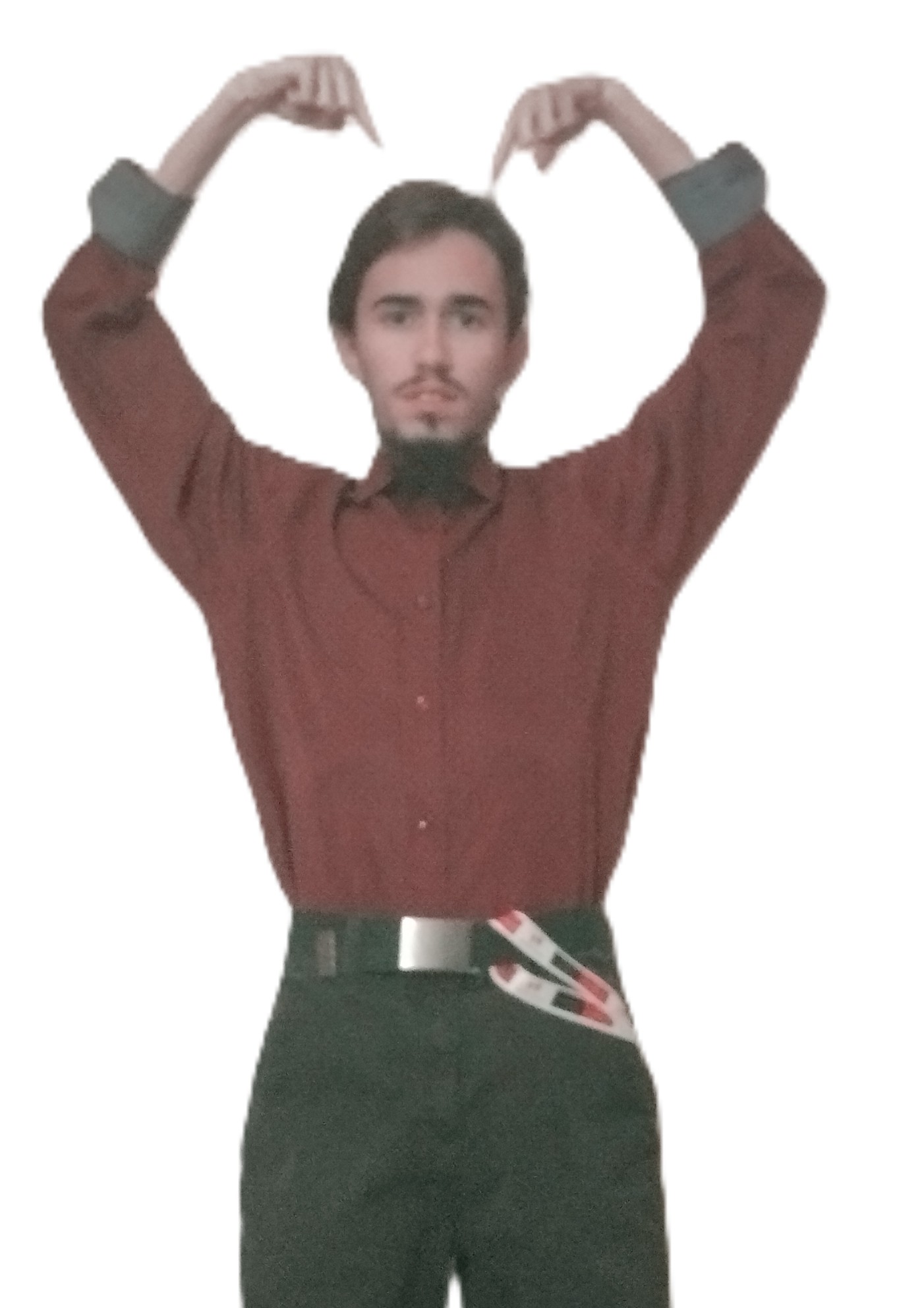}
                \caption{Come to me}
            \end{subfigure}&
            \begin{subfigure}[t]{0.2\textwidth}
                \centering
                \includegraphics[width=\linewidth]{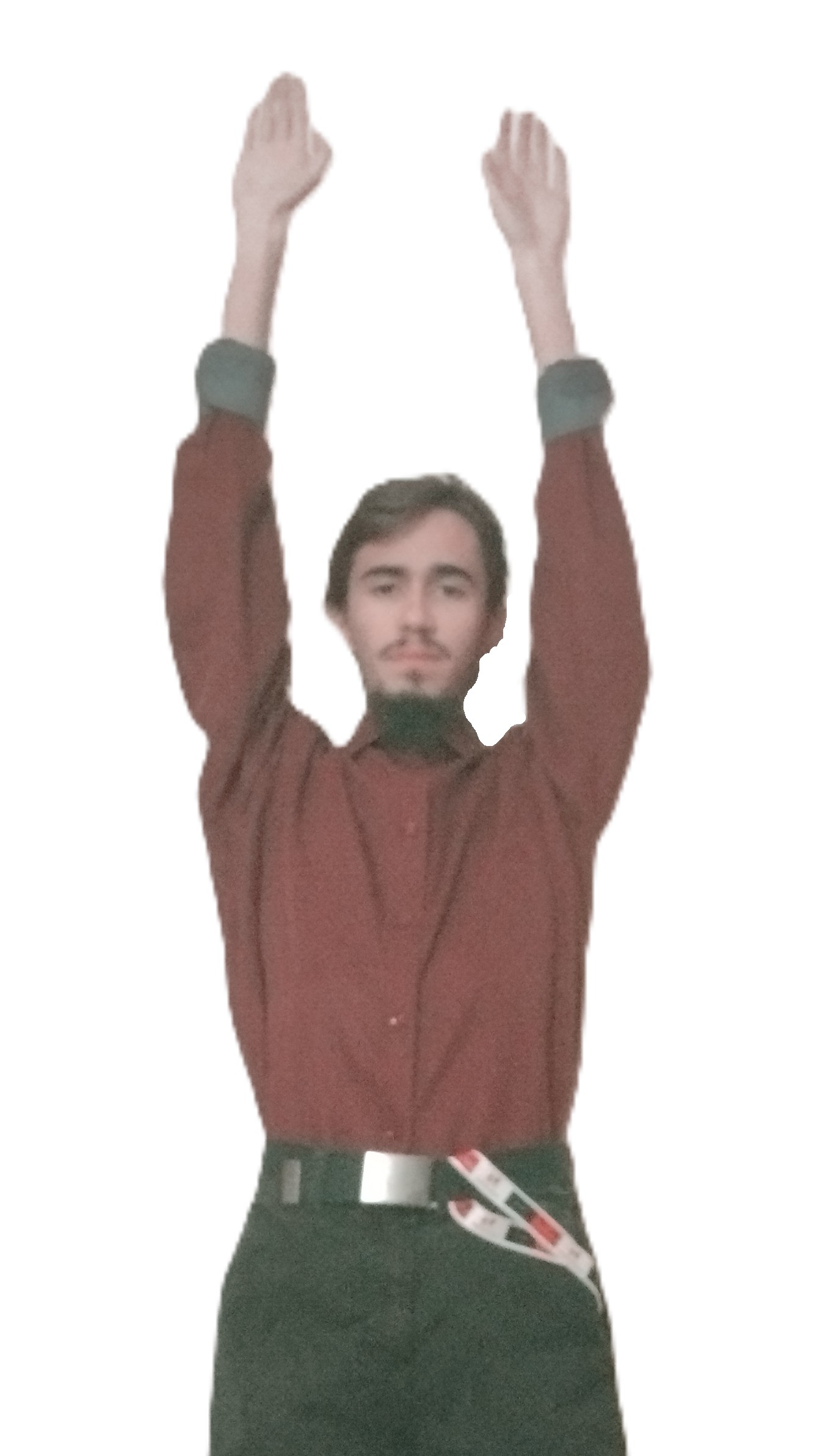}
                \caption{I need help}
            \end{subfigure}&
            \begin{subfigure}[t]{0.2\textwidth}
                \centering
                \includegraphics[width=\linewidth]{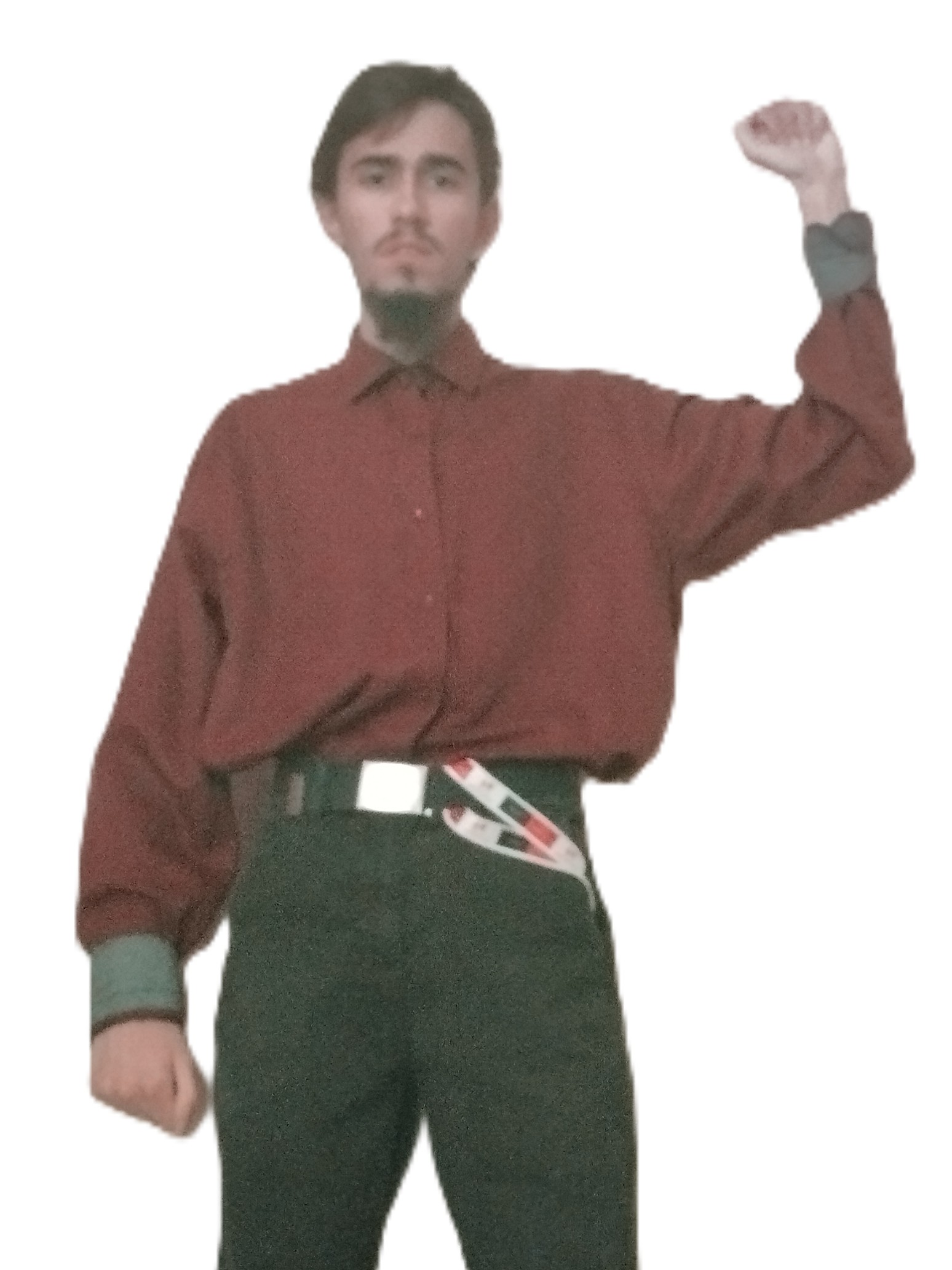}
                \caption{Freeze (stop)}
            \end{subfigure}&
            \begin{subfigure}[t]{0.2\textwidth}
                \centering
                \includegraphics[width=\linewidth]{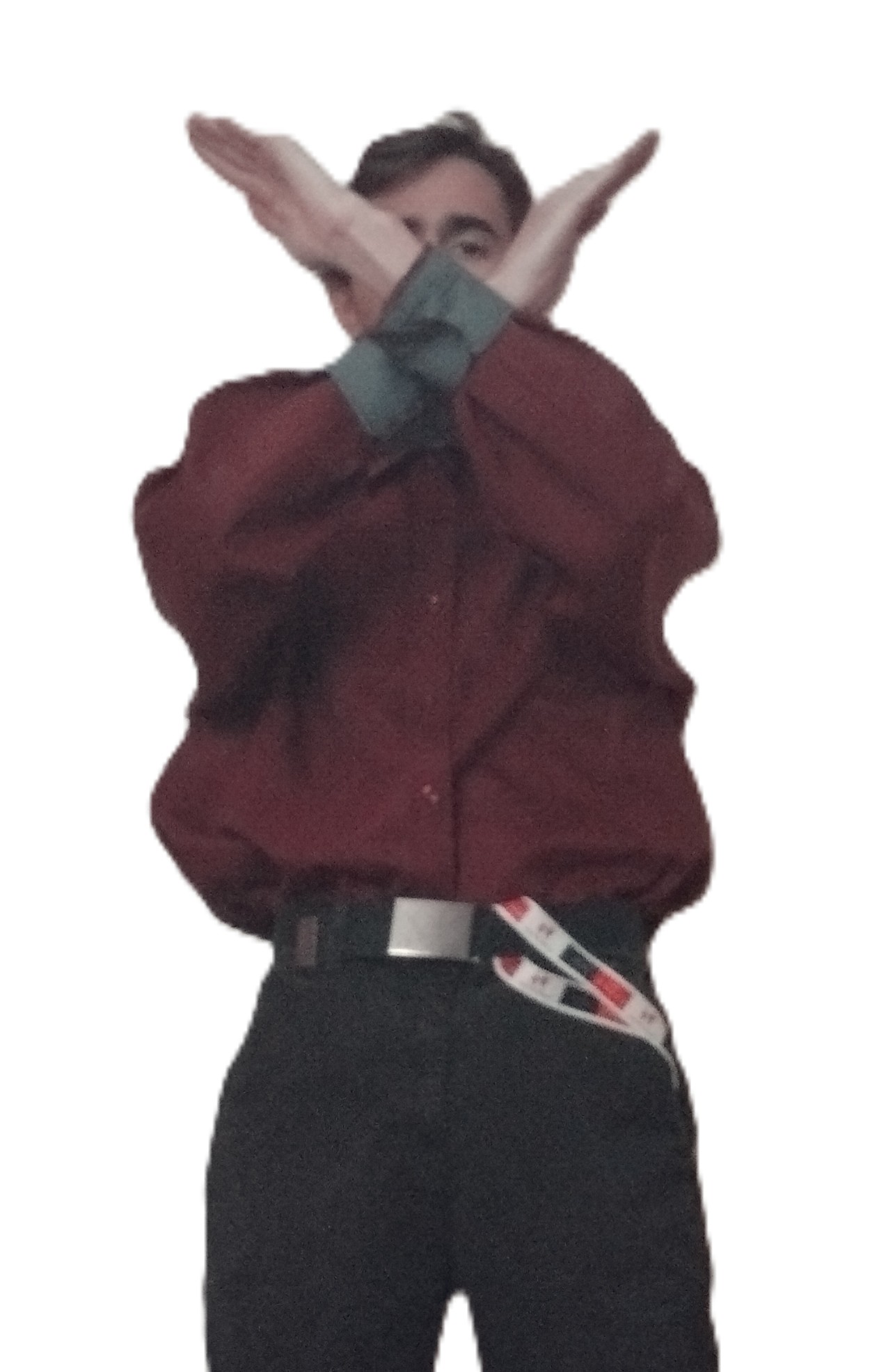}
                \caption{Emergency situation}
            \end{subfigure}
            \\
            
            \begin{subfigure}{0.2\textwidth}
                \centering
                \includegraphics[width=\linewidth]{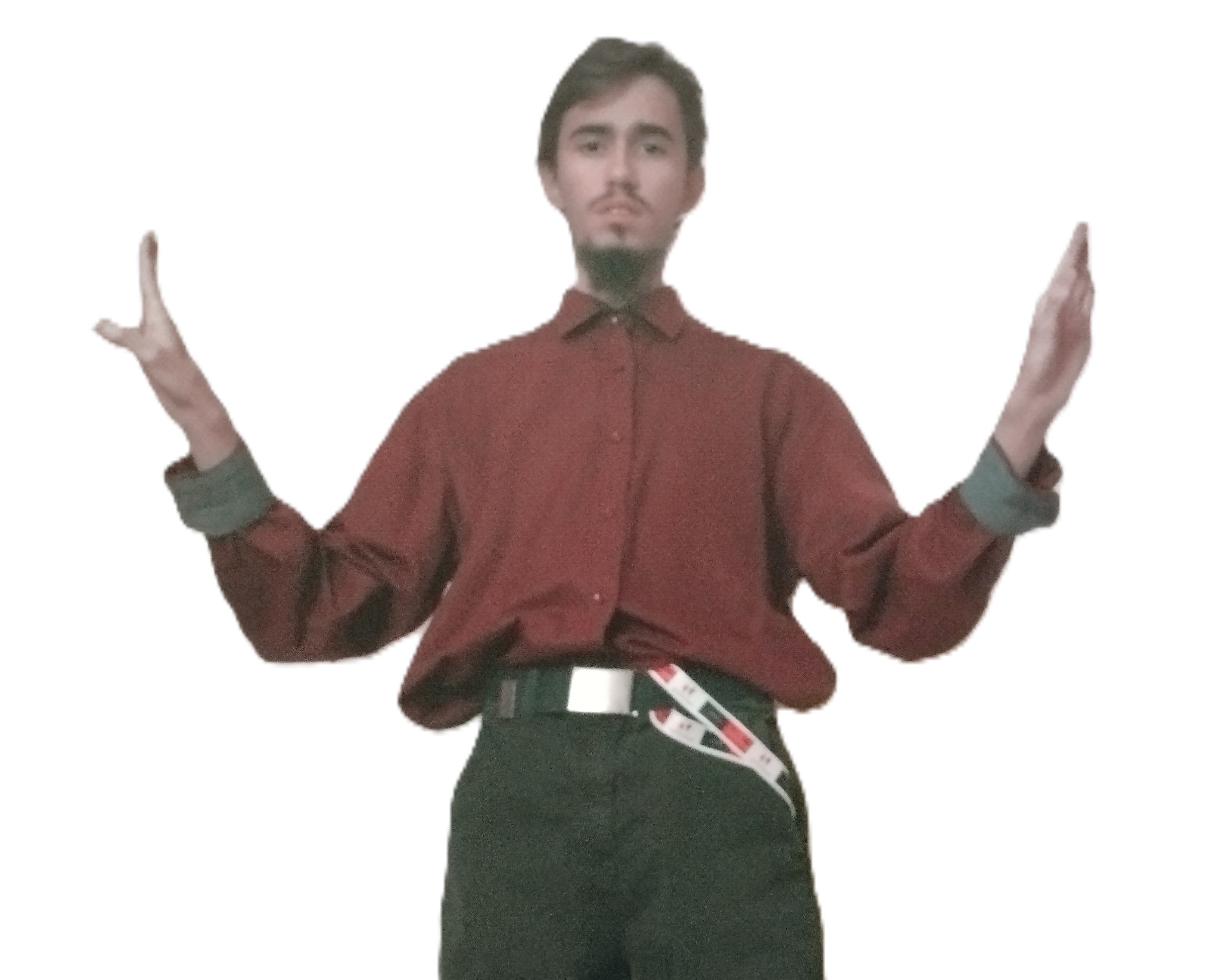}
                \caption{Move away from here}
            \end{subfigure}&
            \begin{subfigure}[t]{0.2\textwidth}
                \centering
                \includegraphics[width=\linewidth]{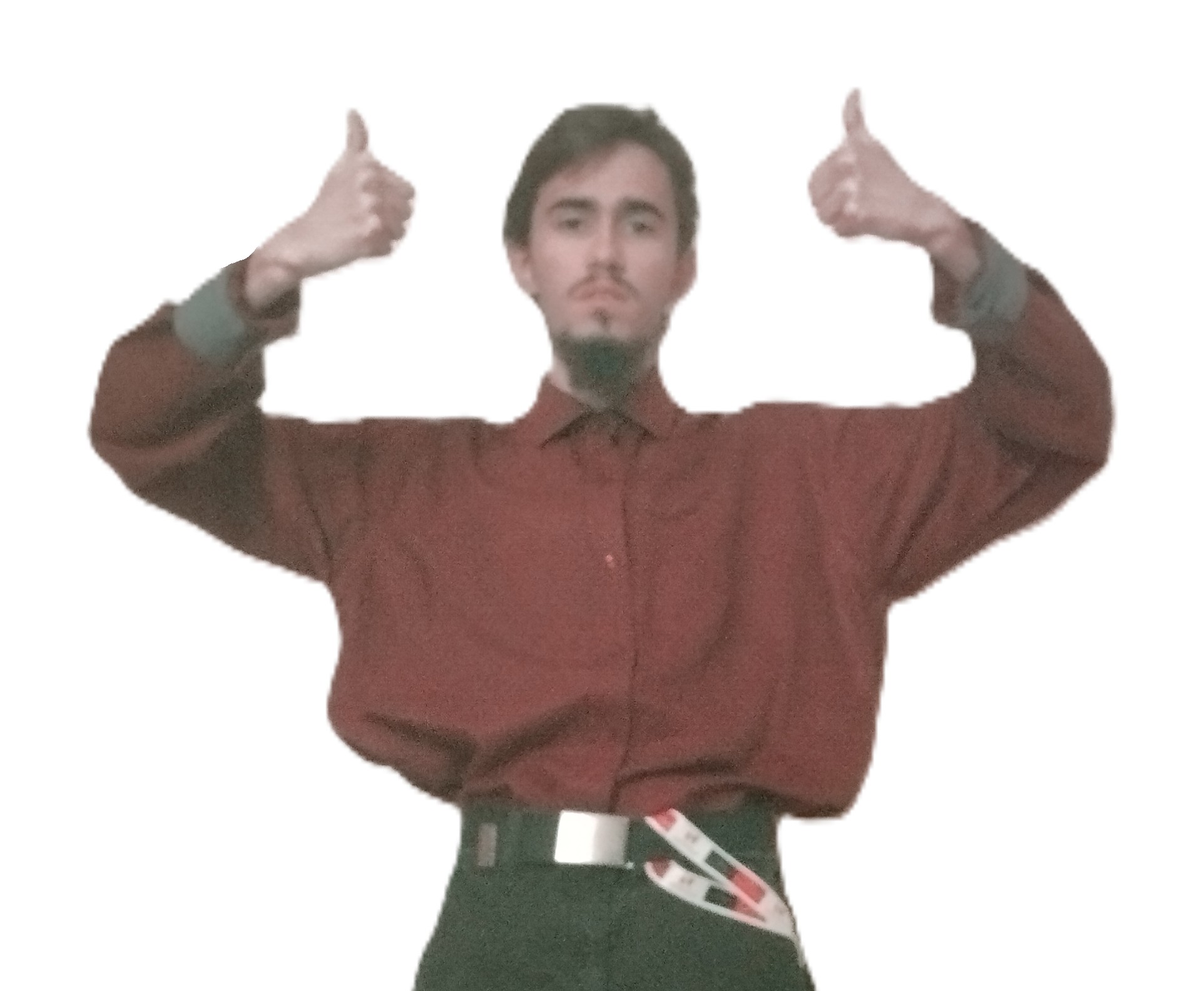}
                \caption{Ok to go}
            \end{subfigure}&
            \begin{subfigure}[t]{0.2\textwidth}
                \centering
                \includegraphics[width=\linewidth]{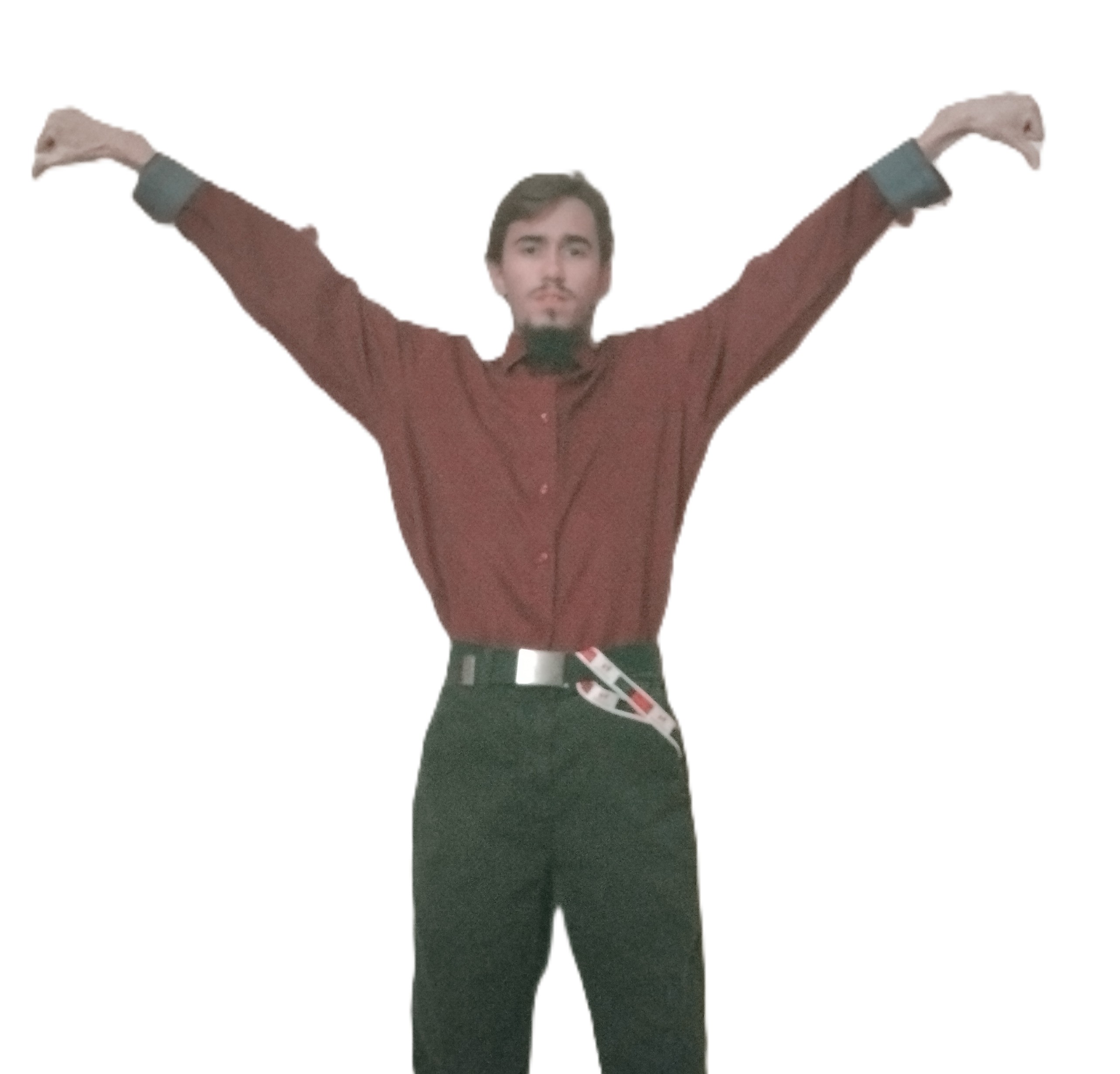}
                \caption{Evacuate the area}
            \end{subfigure}&
            \begin{subfigure}[t]{0.2\textwidth}
                \centering
                \includegraphics[width=\linewidth]{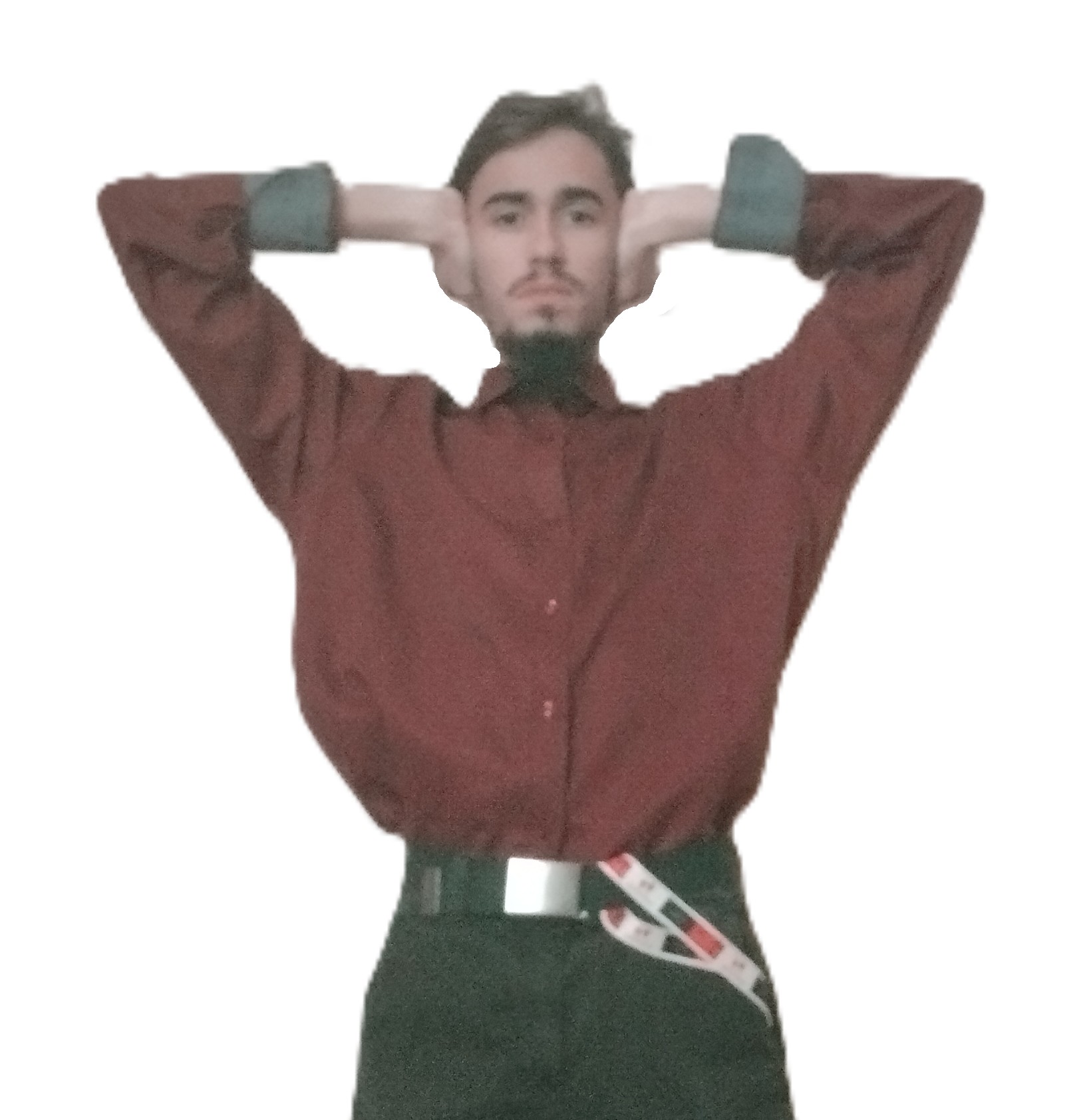}
                \caption{I lost connection}
            \end{subfigure}
            \\
            
            \begin{subfigure}{0.2\textwidth}
                \centering
                \includegraphics[width=\linewidth]{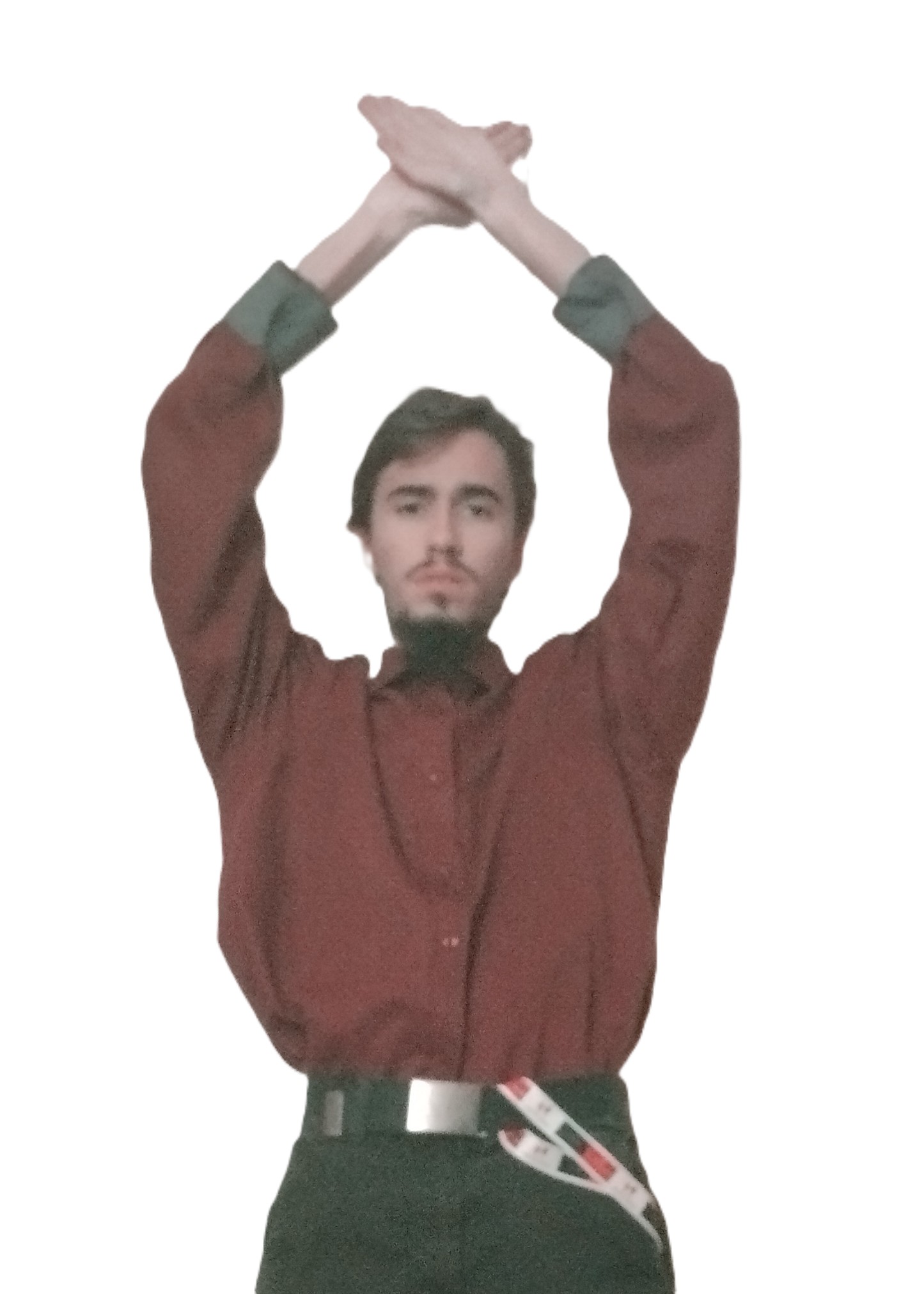}
                \caption{Operation finished}
            \end{subfigure}&
            \begin{subfigure}[t]{0.2\textwidth}
                \centering
                \includegraphics[width=\linewidth]{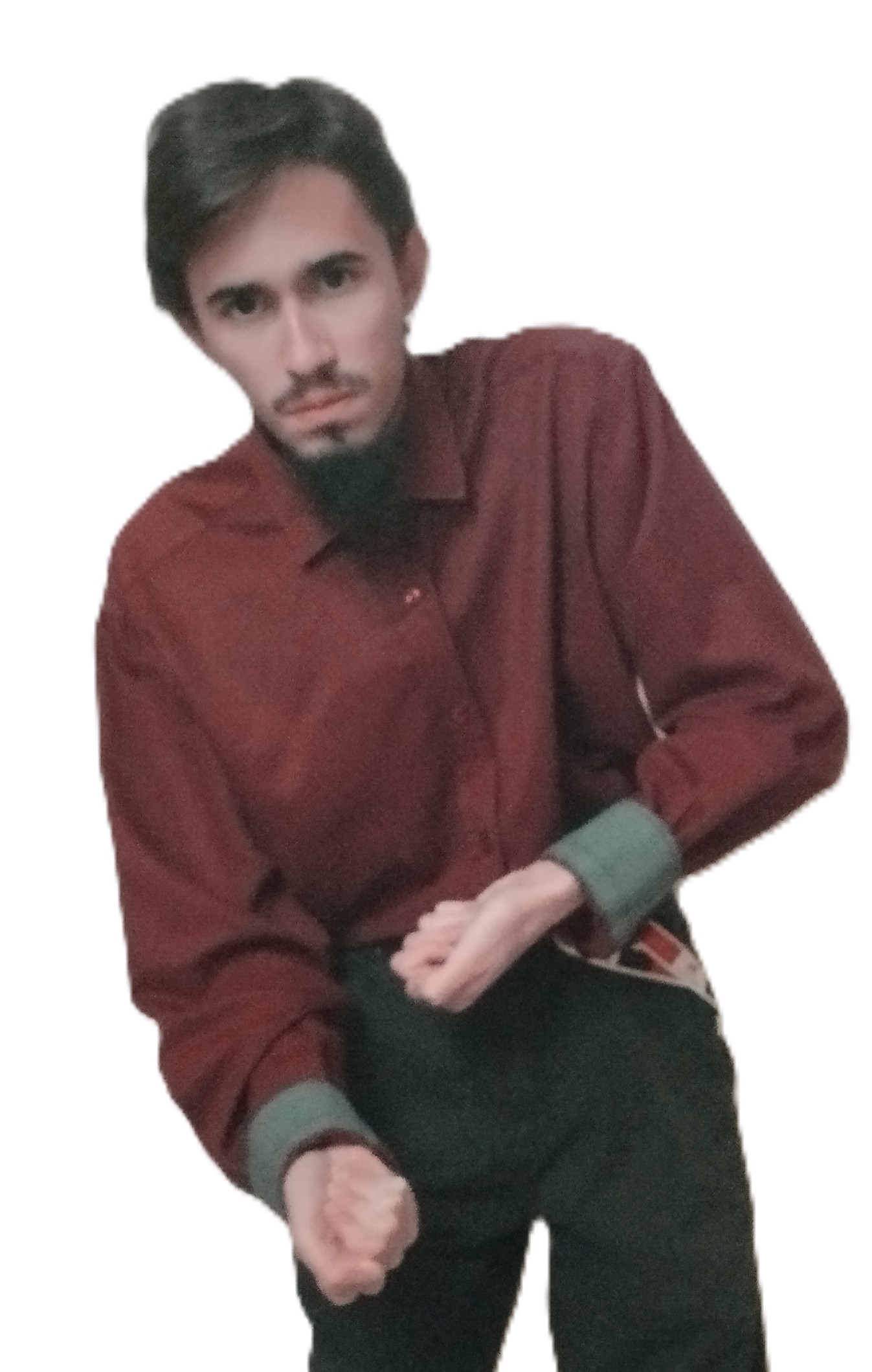}
                \caption{Fetch a shovel}
            \end{subfigure}&
            \begin{subfigure}[t]{0.2\textwidth}
                \centering
                \includegraphics[width=\linewidth]{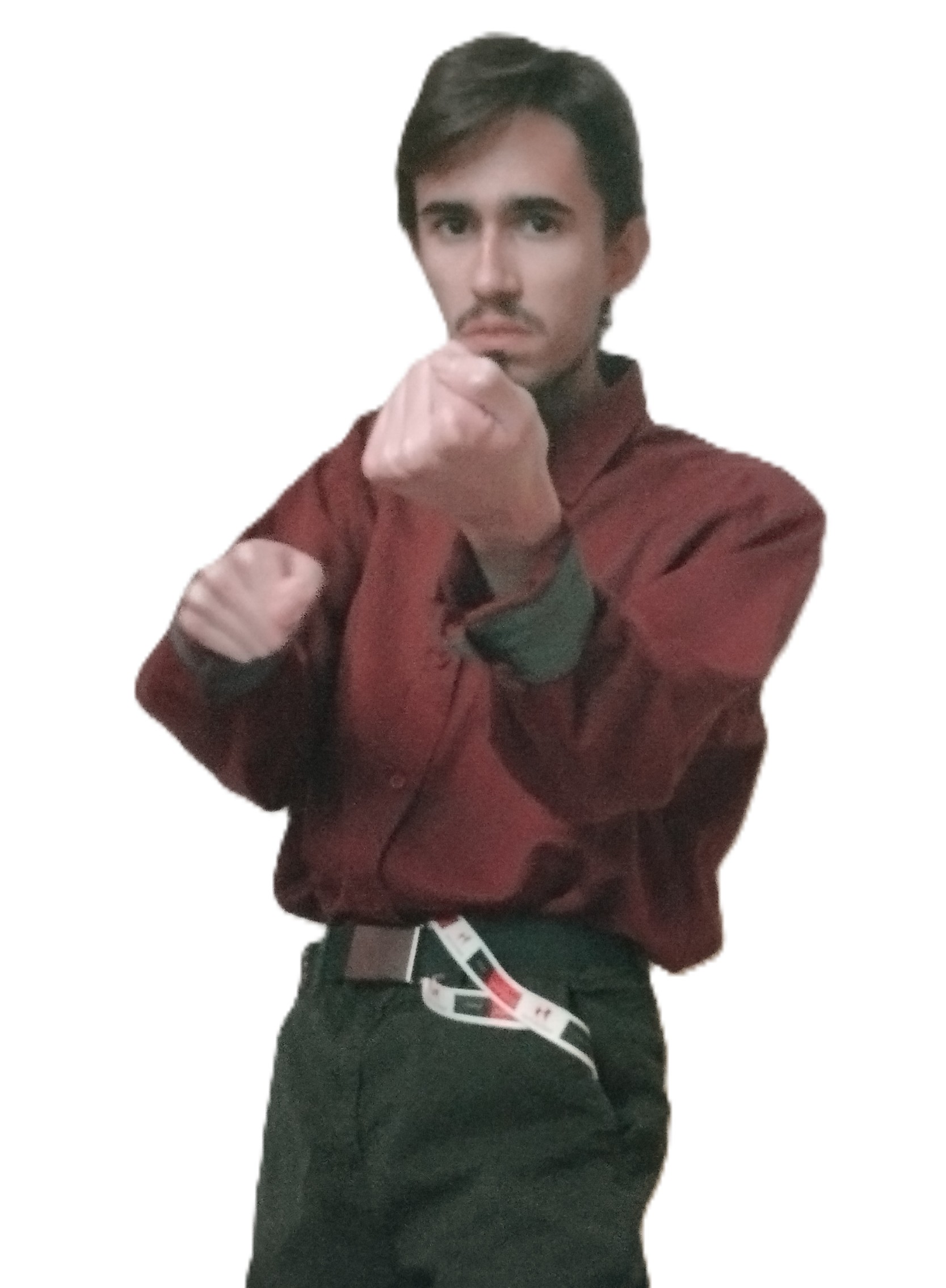}
                \caption{Fetch an ax}
            \end{subfigure}&
            \begin{subfigure}{0.2\textwidth}
                \centering
                \includegraphics[width=\linewidth]{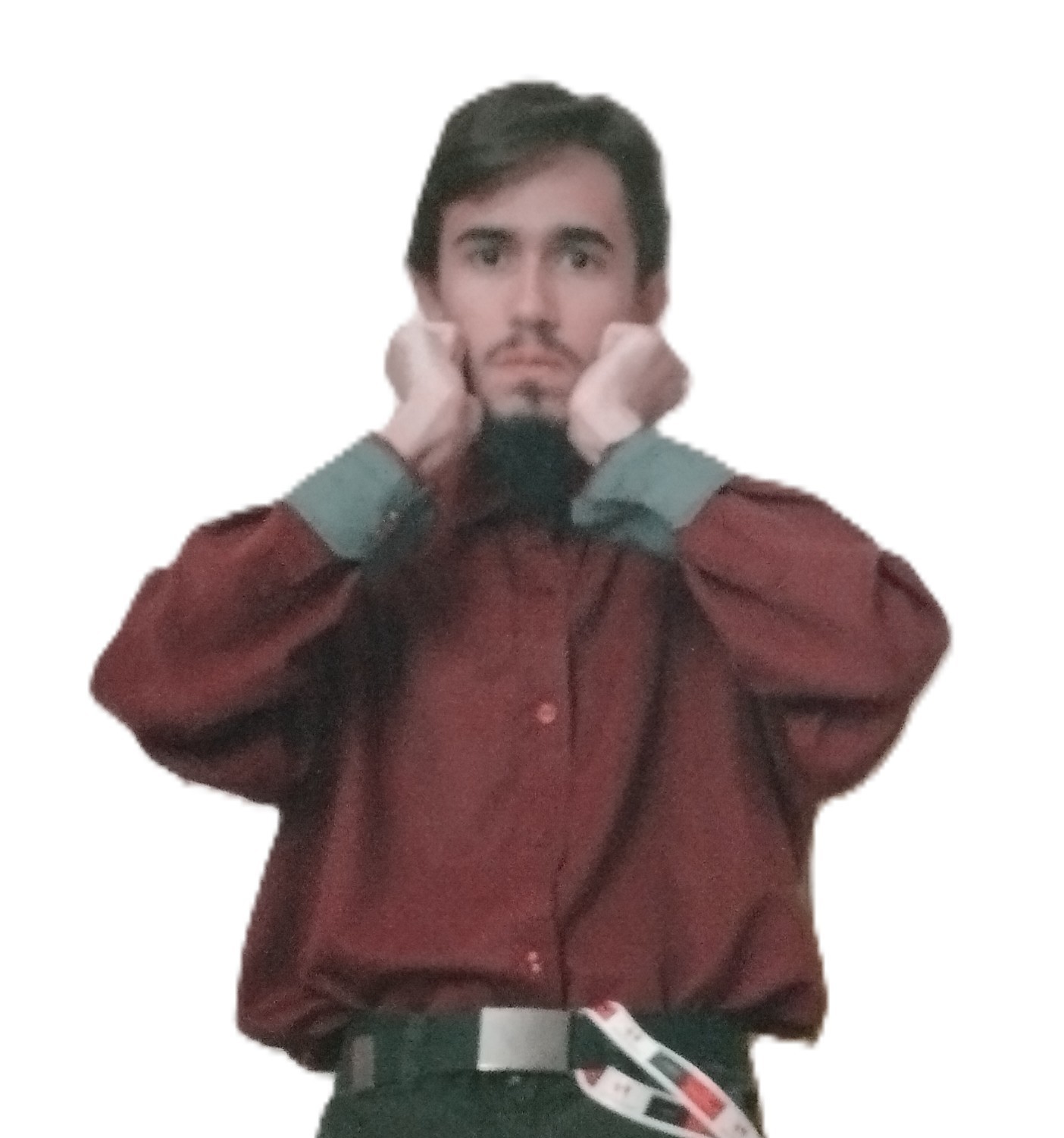}
                \caption{Fetch a gas mask}
            \end{subfigure}
            \\
            
        \end{tabular}
        \caption{Examples of the 12 defined gestures. All of them correspond to specific UGV commands, tailored for first response scenarios.}
        \label{fig:semantic-corpus}
    \end{figure*}

    The proposed dataset contains 12 distinct human gestures (shown in Fig. \ref{fig:semantic-corpus}), which either are custom ones or were obtained by adopting existing hand signals used by task forces and firefighters. Furthermore, this list was presented to experienced FRs and iteratively refined based on their feedback. These gestures correspond to a set of commands to which the robot has to respond. This gesture/command mapping has as follows:
    
    \begin{enumerate}[label=(\alph*)]
        \item \textbf{Come to me:} Indicating the head with both hands. The robot approaches the FR who performs the gesture.
        \item \textbf{I need help:} Hands raised. The FR calls for support (e.g., the FR cannot extinguish a fire alone so needs additional FRs to provide aid). The robot transmits the situation to the command center.
        \item \textbf{Freeze (stop):} Fist at head level, performed with either the left or right hand. The robot stops moving.
        \item \textbf{Emergency situation:} Crossing hands at head or chest level. Performed to indicate that an emergency situation has occurred (e.g., a FR has been trapped in debris).
        \item \textbf{Move away from here:} `Spreading' hands. The robot moves away from its current position (e.g., it was obstructing the passage of vehicles).
        \item \textbf{Ok to go:} Thumbs up with both hands at the height of the head. The robot continues its normal path.
        \item \textbf{Evacuate the area:} Hands stretched at head level with thumbs down. The robot leaves the area due to immediate danger (e.g., a tank full of flammable liquid is about to explode).
        \item \textbf{I lost connection:} Covering the ears with both hands, indicating that the FR is facing communication issues (e.g., wireless communication is not working).
        \item \textbf{Operation finished:} Hands raised with palms crossed above the head. The operation has finished, so the robot returns to the ground station.
        \item \textbf{Fetch a shovel:} Pretending to dig the ground. The robot fetches a shovel.
        \item \textbf{Fetch an ax:} Pretending to cut something. The robot fetches an ax.
        \item \textbf{Fetch a gas mask:} Mimicking gas mask filters with the hands. The robot fetches a gas mask.
    \end{enumerate}

    \subsection{Data gathering protocol}

    The dataset was collected by a total of 7 participants. These subjects were asked to perform each one of the 12 gestures once at 6 or 7 different distances (approximately from 1 to 7 meters), aiming to enhance robustness to recognition range variability \cite{LRHG_9561189,urgr_bamani2024ultra}. This process was repeated in 3 different scenes (1 outdoors and 2 indoors) to make the model invariant to the surrounding environment. Each gesture was captured by two different Intel RealSense D415 cameras, positioned in different heights and arbitrary viewpoint angles to increase diversity, at resolution 480x640 for both RGB and depth modalities and saved in PNG format. Partially occluded samples or frames with motion blur were also considered, regarding related edge cases of the in-the-wild deployment. The data curation was facilitated by a script, which was showing sequentially the gesture that the subject should perform and automatically captured the frames after a timelapse. CSVs files were used to keep track of the process, saving the status of each sample (whether it has been captured or not) and necessary information (e.g. subject identifier, paths to RGBD frames, distance number). The dataset was examined by a supervisor to remove duplicated or incorrect samples.
    
    \subsection{Dataset statistics}

    Gestures are equally distributed within the dataset, aiming to prevent issues that may occur due to class imbalance. Per-scene and per-subject distributions are shown in the following figures (Fig. \ref{fig:scene-dist} and Fig. \ref{fig:subject-dist}). Scene2 and scene3 are indoor environments while scene1 is outdoors. The dataset consists of 3312 RGBD pairs in total.

    \begin{figure}
        \centering
        \includegraphics[width=0.8\linewidth]{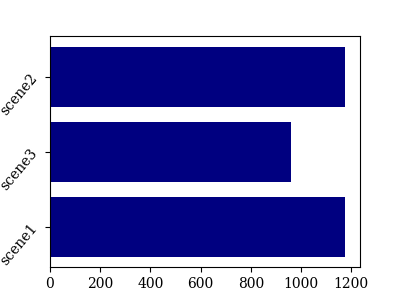}
        \caption{Distribution of scenes.}
        \label{fig:scene-dist}
    \end{figure}

    \begin{figure}
        \centering
        \includegraphics[width=0.8\linewidth]{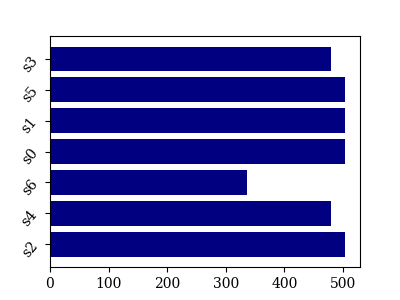}
        \caption{Distribution of subjects.}
        \label{fig:subject-dist}
    \end{figure}

\section{Employed Methods}
    \label{sec:methods}

    In this section, we discuss in detail the methods that we have adopted, following \cite{kapitanov2024hagrid,pipelines_dang2023lightweight}. Newer image classifiers considered to have very large capacity compared to the complexity of the task, thus they may suffer from overfitting, so they are left for future work.
    
    \subsection{ResNet} 
    
    Most of the existing works on CSLR (the 2DCNN+1DCNN+BiLSTM pipelines \cite{vac_cslr_min2021visual}) still use ResNet \cite{resnet_he2016deep} to obtain per-frame representations, due to its ability to efficiently propagate the gradients, through residual connections, facilitating the optimization process. In more detail, the authors of ResNet proposed five different settings, corresponding to various scales (with 18, 34, 50, 101 and 152 layers). All of these architectures share the same building block, which is the core of the publication; the residual (aka shortcut) connections, which aim to mitigate optimization difficulties and in particular the fact that deep architectures tend to be more difficult to be trained. The residual blocks incorporate an identical mapping, which is added to the output. Thus, the resulting module can be formulated as $y=F(x,\{W_i\})$ where $\{W_i\}$ are the weights of the in between layers. If the dimensions do not match, linear projection is applied and the formula becomes $y=F(x,\{W_i\})+W_sx$. We chose one small (ResNet-18) and one intermediate (ResNet-50) configuration as larger models are more prone to overfitting (regarding the limited size of the dataset) and require more computational resources, restricting real-world applicability.

    \subsection{ResNeXt}
    
    ResNeXt \cite{resnext_xie2017aggregated} is an extension of the ResNet, which aims to increase the capacity of the former with a negligible computational overhead. The core idea lies in modifying the residual block, through transforming it into a multi-branch module. The number of these branches, called “cardinality”, is revealed as a new hyperparameter together with width (e.g. number of channels) and depth (e.g. number of layers). The outputs of each stream are aggregated via addition. We selected the smallest method provided by PyTorch \cite{pytorch_paszke2019pytorch}, the ResNeXt-50 32x4 (i.e. 50 layers with cardinality equal to 32 and width equal to 4).
    
    \subsection{EfficientNet} 
    
    The authors of EfficientNet \cite{efficient_net_tan2019efficientnet} targeted to efficiently scale DL-based image classifiers, balancing computational complexity and accuracy. In particular, they made two observations that drove them to introduce the compound scaling scheme: 1) tuning width, depth and resolution of the model independently increases the accuracy, but the gain saturates and 2) in order to develop both a fast and accurate method, these values should be changed simultaneously. The compound scaling framework relies on the compound coefficient, which is employed to modify the architecture following equations $d=a^{\phi}, w={\beta}^{\phi}, {\gamma}^{\phi}$, where $d, w, r$ correspond to depth, width and resolution, respectively. The $\alpha, \beta, \gamma$ constants are determined through grid search, which control how the available computational resources (modeled by $\phi$) are distributed within the network. This approach was applied to existing architectures (e.g. ResNet \cite{resnet_he2016deep}) and to a custom one, termed EfficientNet, which was obtained through multi-objective NAS, optimizing accuracy and FLOPs simultaneously. The low computational cost (5.2M parameters and 0.39 GFLOPs) makes it suitable for resource-constrained deployment (e.g. running onboard on the UGV).



\section{Experimental protocols and results}
    \label{sec:experiments}

    In this section, we discuss the experimental setup for evaluating the performance of the various methods in our dataset. The first subsection analyzes the training details, the second introduces the general protocol and the third proposes a subject-independent setting for measuring the generalization ability to unseen signers. All the experiments concern RGB-based methods, following the trends on static gesture recognition.

    \subsection{Training setup}

    Regarding the small size of the dataset, the large-scale version of HaGRID \cite{kapitanov2024hagrid} is utilized for pretraining the RGB-based methods. The batch size was set to 32 for all models. AdamW was selected for optimization due to its ability to mitigate overfitting, with initial learning rate 0.01 and weight decay 0.0001. The learning rate is multiplied by 0.9 after each epoch. All models were trained for 100 epochs using PyTorch \cite{pytorch_paszke2019pytorch} on a NVIDIA GeForce RTX 3080 Ti GPU. Applying augmentation (e.g. random rotation) was examined but we found that it downgraded the performance. F1-Score was adopted as a classification metric, following \cite{kapitanov2024hagrid}. We choose for testing the weights that maximize the metric value in the validation set.

    \subsection{Uniform protocol}

    The dataset is randomly split into train, val, test sets with 2304, 504, 504 samples, respectively, such that the classes are uniformly distributed within each split to avoid issues due to class imbalance. The experimental results are shown in the following table (Table \ref{tab:subject-dependent-results}). Pretraining the methods on HaGRID was found to be essential for mitigating overfitting. EfficientNet achieves superior performance, probably because its small size make it less vulnerable to overfitting leading to better generalization.

    \begin{table}[]
        \centering
        \caption{Reported F1-Score (in percentage) for the uniform protocol.}
        \begin{tabular}{cccccc}
            \toprule
            \textbf{Method} & \textbf{Modality} & \textbf{Pretrained} & \textbf{Train} & \textbf{Val} & \textbf{Test} \\
            \midrule
            
            ResNet-18 & RGB & HaGRID & 99.78 & 92.26 & 92.46 \\
            ResNet-50 & RGB & HaGRID & 99.01 & 90.27 & 90.07 \\
            ResNeXt-50 & RGB & HaGRID & 100.0 & 94.24 & 90.87 \\
            EfficientNet-B0 & RGB & HaGRID & 99.86 & 97.22 & 96.42 \\

            \bottomrule
        \end{tabular}
        \label{tab:subject-dependent-results}
    \end{table}
    
    \subsection{Subject-independent protocol}
    
    Aiming to evaluate the robustness of models trained on our dataset, we defined a subject-independent protocol, that is, gestures of each unique signer are included only in a specific split; thus, the model is asked to recognize gestures performed by unseen subjects in the validation and test splits. Therefore, the experimental results are indicative of the generalization ability of the models and the need for curating more samples or not. The dataset is split into 2304 samples (corresponding to 5 subjects) for training, 504 for validation (1 subject) and 504 for testing (1 subject). The experimental results are shown in the table below (Table \ref{tab:subject-independent-results}). The metric is decreased significantly compared to the results for the uniform protocol, as the limited number of subjects participating in the dataset leads to insufficient generalization ability. Yet, EfficientNet still achieves superior performance.

    \begin{table}[]
        \centering
        \caption{Reported F1-Score (in percentage) for the subject independent protocol. The performance is seriously affected when the models are enforced to recognize gestures performed by unseen signers.}
        \begin{tabular}{cccccc}
            \toprule
            \textbf{Method} & \textbf{Modality} & \textbf{Pretrained} & \textbf{Train} & \textbf{Val} & \textbf{Test} \\
            \midrule
            
            ResNet-18 & RGB & HaGRID & 99.95 & 88.88 & 65.67 \\
            ResNet-50 & RGB & HaGRID & 94.53 & 75.79 & 52.18 \\
            ResNeXt-50 & RGB & HaGRID & 98.56 & 57.93 & 39.68 \\
            EfficientNet-B0 & RGB & HaGRID & 99.86 & 95.43 & 87.73 \\
            
            \bottomrule
        \end{tabular}
        \label{tab:subject-independent-results}
    \end{table}

\section{Limitations}
    \label{sec:limitations}

    Our dataset suffers from several limitations, which could be mitigated in a future work in order to increase robustness to in-the-wild deployment:
    \begin{itemize}
        \item \textbf{Laboratory samples:} The dataset has been collected by students wearing arbitrary casual clothing. This may not be optimal for recognizing gestures performed by FRs, who may wear specific uniforms and personal protective equipment (e.g., gloves, helmets and masks). Furthermore, the samples have been recorded in a university, which also differs a lot from the post-disaster environments on which FRs mainly operate. However, these limitations can be mitigated by increasing the dataset size and diversity, enforcing the models to pay attention only in the most discriminative regions of the image (e.g. shape of the hands), discarding redundant information. Diffusion models for data augmentation could also be examined \cite{alimisis2025advances}.
        \item \textbf{Gender and race imbalance:} Only 2 out of the 7 subjects are women, while all participants are white. This may pose significant barriers for accurate recognition in real-world scenarios.
    \end{itemize}
    
\section{Conclusion}
    \label{sec:conclusion}

    Considering the limitations of existing datasets for gesture-based HRI and the absence of a dataset tailored for use by FRs, in this paper we introduce a novel set of gestures, mapped to robot commands, and we present the corresponding RGBD dataset, collected at multiple distances and viewpoints to increase diversity. Further, we perform some baseline experiments using well-established image classifiers considering two evaluation protocols (subject dependent and subject independent). Finally, we end up with the limitations of our study.

\section*{Acknowledgment}

    The authors would like to thank members of the HUA Computer Vision Group for participating in the data collection process.
 
\balance

\bibliography{references}
\bibliographystyle{plain}

\end{document}